\documentclass{article}
\usepackage{spconf,amsmath,graphicx}
\usepackage{cite}
\usepackage{amsmath,amssymb,amsfonts}
\usepackage{algorithmic}
\usepackage{graphicx}
\usepackage{textcomp}
\usepackage{xcolor}
\usepackage{academicons}
\usepackage{svg}

\usepackage{subcaption}
\usepackage{autobreak}
\usepackage{subcaption}
\usepackage{bm}
\usepackage{xspace}
\usepackage{mathtools}

\usepackage{times}
\usepackage{epsfig}
\usepackage{booktabs}

\usepackage{amsmath,amssymb,amsfonts}
\usepackage{algorithmic}
\usepackage{graphicx}
\usepackage{textcomp}

\usepackage{framed,multirow}

\usepackage{amssymb}
\usepackage{latexsym}

\usepackage{url}
\usepackage{graphicx}
\usepackage{amsmath}
\usepackage{amssymb}
\usepackage{booktabs}
\usepackage{multicol}
\usepackage{makecell}
\usepackage{multirow}
\usepackage{subcaption}
\usepackage{blindtext}
\usepackage{cite}
\usepackage{tabularx}
\usepackage{hyperref}
\title{A3S: Adversarial learning of semantic representations for Scene-Text Spotting}
\name{Masato Fujitake}
\address{Fast Accounting Co., Ltd.
 \\
{\tt\small fujitake@fastaccounting.co.jp}
}
\begin{document}
%\ninept
%
\maketitle
\begin{abstract}
    Scene-text spotting is a task that predicts a text area on natural scene images and recognizes its text characters simultaneously.
It has attracted much attention in recent years due to its wide applications.
Existing research has mainly focused on improving text region detection, not text recognition.
Thus, while detection accuracy is improved, the end-to-end accuracy is insufficient.
Texts in natural scene images tend to not be a random string of characters but a meaningful string of characters, a word.
Therefore, we propose adversarial learning of semantic representations for scene text spotting (A3S) to improve end-to-end accuracy, including text recognition.
A3S simultaneously predicts semantic features in the detected text area instead of only performing text recognition based on existing visual features.
Experimental results on publicly available datasets show that the proposed method achieves better accuracy than other methods.
\end{abstract}
\section{Introduction}
\label{sec:intro}
Scene-text spotting is a task for detecting and recognizing texts in natural scene images.
It has recently attracted attention for its usefulness in real-world applications such as document understanding and automated driving~\cite{zhang2022testr, qiao2021mango, liao2020masktextspotterv3, fujitake2021tcbam}.
The task consists of two main processes: first, to predict the coordinate positions of text areas in natural scene images, and second, to recognize texts in the detected regions.
Accurate detection of text regions in natural scene images is difficult due to various variations, such as curved text strings and complex layouts.
For this reason, many works have been proposed in recent years that have greatly improved the accuracy of text region detection~\cite{liao2020masktextspotterv3, liu2021abcnetv2, feng2019textdragon}.
However, end-to-end accuracy has not been sufficiently improved due to errors in text recognition.
To recognize text in scene images, most of the methods proposed in recent years rely on only visual information in the detected text area.
However, such an approach is vulnerable to change in various fonts, styles, colors, and shapes, which leads to misspelled recognized texts.

In natural scene images, many texts have semantic information like meaningful words and sentences, and random character strings are rare.
Therefore, we considered using semantic information from language models can help to reduce text recognition errors.
In this paper, we propose a novel method, \textbf{A}dversarial learning of \textbf{S}emantic representations for \textbf{S}cene-text \textbf{S}potting, A3S.
In text recognition, A3S simultaneously predicts the text area's semantic features instead of directly predicting text from the visual features obtained from images.
The predicted semantic features of the text are trained to match the ones of a pre-trained language model.
Similar studies that utilize visual and semantic features have densely matched in the same space~\cite{rodriguez2013label}. 
Still, since these features are strictly different, we propose to match them flexibly through adversarial learning.
This approach improves end-to-end accuracy by predicting features that consider semantic information without relying excessively on visual information.

Our method improves the end-to-end accuracy in scene-text spotting. 
Furthermore, it achieves state-of-the-art accuracy on several public datasets.
In summary, the contributions of this paper are summarized as follows:

\begin{itemize}
    \item We propose A3S, an adversarial learning method for scene text spotting.
 To the best of our knowledge, this is the first work that jointly leverages semantic representations for scene-text spotting.
    \item With a simple but effective approach, we improve 6.9\% accuracy on the CTW1500 dataset.
We also archive state-of-the-art accuracy on several benchmarks.

\end{itemize}

\begin{figure*}[htp]
	\centering
	\includegraphics[width=1.50\columnwidth, keepaspectratio]{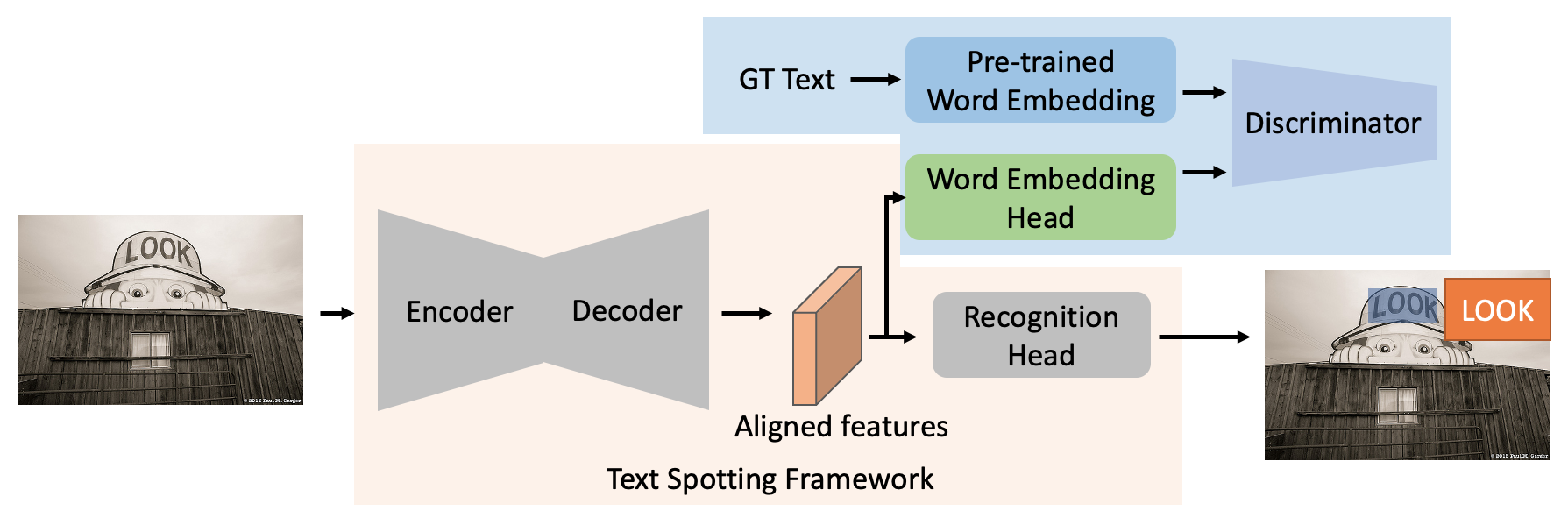}
	\caption{
The architecture of the proposed method, adversarial learning of semantic representations for text spotting (A3S).
It consists of the text spotting framework~\cite{liu2021abcnetv2}, the word embedding head, the pre-trained word embedding, and the discriminator.
The word embedding head predicts the semantic features from the aligned detection candidate features obtained inside the text spotting framework.
The discriminator optimizes to close the predicted and semantic features from the pre-trained word embedding.
	}
	\label{fig:overall_architecture}
    \vspace*{-1.00\baselineskip}
\end{figure*}

\section{Related Works} \noindent
\textbf{Scene-Text Spotting.}
Scene-text spotting requires text detection~\cite{zhou2017east} and recognition~\cite{shi2016crnn} simultaneously.
Two-stage approaches~\cite{liao2017textboxes} are proposed, which individually develop detection and recognition modules and join them during inference.
However, multi-steps may require detailed tuning, leading to sub-optimal performance and time consumption.
On the other hand, recent literature focuses on end-to-end methods~\cite{li2017towards, liu2021abcnetv2, zhang2022testr, yi2021pnats}, which train both modules simultaneously.
Some methods~\cite{liu2018fots, feng2019textdragon} proposed special RoI operations to sample features in the text area.
TESTER~\cite{zhang2022testr} proposes query-based decoders to remove complex RoI operations.
Mask TextSpotter series detect and recognize text instances of arbitrary shapes by segmenting the text regions~\cite{liao2020masktextspotterv3}. 
ABCNet series~\cite{liu2021abcnetv2} introduces parametric Bezier curve representations for curved texts.
SwinTextSpotter~\cite{huang2022swintextspotter} proposes a recognition conversion mechanism to explicitly guide text localization.
Although they improve detection accuracy by focusing on detecting arbitrary-shaped text areas, the end-to-end accuracy is insufficient.
In contrast, our work focuses on improving text recognition on scene-text spotting.
Although GLASS~\cite{ronen2022glass} proposed an attention mechanism recently to fuse visual global and local information for text recognition accuracy, it only relies on vulnerable visual features.
Our method exploits the semantic representations of the text to improve recognition.

\noindent
\textbf{Visual-Semantic embedding.}
Semantic information provides different information from the visual one obtained from images.
Therefore, some studies proposed using both representations simultaneously.
Methods have been proposed for embedding them in the same Euclidean space~\cite{rodriguez2013label} or estimating one feature from the other and fusing them together~\cite{cheng2021vdsl}.
However, they utilize different representations simultaneously and directly, which may interfere with each other.
In contrast, we enable flexible learning by introducing adversarial learning.

\label{sec:relatedwork}

\section{Method} \label{sec:method}
The overall architecture of A3S is presented in Fig~\ref{fig:overall_architecture}, which consists of four components: (1) a baseline text spotting framework~\cite{liu2021abcnetv2}; (2) word embedding head; (3) pre-trained word embedding; and (4) a discriminator for adversarial learning.
In the following, we briefly explain the text spotting framework and introduce the details of the proposed modules and optimization.

\subsection{Text Spotting Framework}
We employ ABCNet v2~\cite{liu2021abcnetv2} as the baseline text spotting framework, as shown in the orange-colored area in Fig~\ref{fig:overall_architecture}.
It utilizes a single-shot, anchor-free convolutional neural network as the detection framework and the lightweight attention mechanism for recognition.
Based on the predicted detection results, the features are passed to the text recognition head through BezierAlign~\cite{liu2021abcnetv2} to precisely align the visual features of each arbitrarily shaped text area.
The framework adapts end-to-end optimization for detection and recognition.

\subsection{Word Embedding Head}
We propose a word embedding head to enable feature prediction that considers semantic information in a text-spotting framework.
It is a simple neural network shown in Table~\ref{tab:word_embedding_structure}  that estimates semantic features from visual features obtained by detection. 
It is optimized through adversarial learning to close the output of the features by the pre-trained word embedding.
Word Embedding Head and Recognition Head share the detected visual features to enable text recognition with semantic representation.
We set the output dimension of the head identical to the pre-trained word embedding.

\begin{table}[bt]
\centering
\caption{
	Structure of Word Embedding Head. 
	For all convolutional layers, the padding size is restricted to 1.
	$n$, $c$, $h$, and $w$ represent the batch size, the channel size, and the height and width of the outputted features, respectively.
	$d$ means the output dimension size of the pre-trained word embedding.
}
\label{tab:word_embedding_structure}
\resizebox{0.9\columnwidth}{!}{%0
\begin{tabular}{c|c|c}
\toprule
\multirow{2}{*}{Layers} & Parameters & Output Size\\ 
&(kernel size, stride) & ($n$, $c$, $h$, $w$)\\ \midrule
conv. layers w/ ReLU $\times$ 2 & (3,1)    & ($n$, 256, $h$, $w$) \\
average pool for $h$  &  $-$    & ($n$, 256, 1, $w$) \\
fc layers w/ ReLU $\times$ 2  &  $-$    & ($n$, $d$)\\
\bottomrule
\end{tabular}
}
    \vspace*{-1.00\baselineskip}
\end{table}

\subsection{Pre-trained Word Embedding}
To generate the ground truth semantic information of text in images, we utilize a pre-trained language model, BERT~\cite{devlin2019bert}.
It is a transformer-based language model~\cite{vaswani2017transformer}, which leverages an attention mechanism that learns contextual relations between words in a text.
The ground truth text to be detected is encoded through BERT, which outputs a semantic vector to the subsequent discriminator.
We use the last hidden states of BERT as the semantic vector.
If the text has multiple words, the output vector is mean pooled.
The pre-trained word embedding's weights are fixed.

\subsection{Adversarial learning}
By adversarial learning~\cite{goodfellow2020generative}, we optimize the word embedding head with the predicted semantic features and the ones from pre-trained language models.
We train the head by training a discriminator that identifies whether features are the predicted ones or the ones from pre-trained word embedding.
A two-class classification using binary cross-entropy loss function is performed with the predicted features as 0 and the ones of the pre-trained word embedding as 1.
A two-class classifier with three fully-connected layers was used as the discriminator.

\subsection{Optimization}
The training method follows the text spotting framework~\cite{liu2021abcnetv2}, except for adversarial learning.
We prepare semantic features as ground truth from the ground truth text in images to train the word embedding head and the discriminator.
We generate features with all the semantic vector elements equal to zero for false-positive detections.
It is then optimized according to the adversarial learning procedure~\cite{goodfellow2020generative}.

The whole loss function consists of three parts, which are defined as follows:
\begin{equation}\label{eq:finetune}
    \mathcal{L} = \alpha\mathcal{L}_{\rm det} + \beta\mathcal{L}_{\rm rec} + \gamma\mathcal{L}_{\rm adv},
\end{equation}
where $\mathcal{L}_{\rm det}$ and $\mathcal{L}_{\rm rec}$ are the detection and recognition loss function in text spotting framework~\cite{liu2021abcnetv2}.
The loss function $\mathcal{L}_{\rm adv}$ is the adversarial learning loss~\cite{goodfellow2020generative}.
$\alpha$, $\beta$ and $\gamma$ are the balance weights for $\mathcal{L}_{\rm det}$, $\mathcal{L}_{\rm rec}$ and $\mathcal{L}_{\rm adv}$, respectively.

\section{Experiments} \label{sec:experiments}
\begin{table}[t]
  \caption{
 End-to-End text spotting performance comparison on CTW1500, ICDAR 2015, and Total-Text datasets.
Following the standard evaluation protocol, we report the end-to-end results over two lexicons: ``None'' and ``Full'' on CTW1500 and Total-Text.
``None'' means that no lexicons are provided, and the ``Full'' lexicon provides all words in the test set.
In ICDAR1500, ``S'', ``W'', and ``G'' mean recognition with the strong, weak, and generic lexicon, respectively.
  }
\centering
  \label{tab:method_overall_result}
\resizebox{1.0\columnwidth}{!}{%
\begin{tabular}{c|cc|ccc|cc}
\toprule
\multirow{2}{*}{Methods} & \multicolumn{2}{c|}{CTW1500} & \multicolumn{3}{c|}{ICDAR2015} & \multicolumn{2}{c}{Total-Text} \\
                         & None         & Full          & S       & W        & G         & None           & Full            \\\midrule 
                         
Text Dragon~\cite{feng2019textdragon} &    38.7     &  72.4     &   82.5     &   78.3  &  65.1  &   48.8   & 74.8       \\
Mask TextSpotter v3~\cite{liao2020masktextspotterv3} &    $-$     &  $-$     &   83.3     &   78.1  &  74.2  &   71.2    &  78.4       \\
PGNet~\cite{wang2021pgnet} &    $-$     &  $-$     &   83.3     &   78.3   &  63.5   &   63.1     &  $-$       \\
ABCNet v2~\cite{liu2021abcnetv2}        &   57.5      &    77.2   &   82.7   &   78.5  & 73.0     &  70.4        &   78.1        \\
TESTR~\cite{zhang2022testr}             &   56.0      &    81.5   &   \textbf{85.2}   &   79.4  & 73.6     &  73.3        &   83.9       \\
SwinTextSpotter~\cite{huang2022swintextspotter}    &  51.8       &    77.0   &   83.9   &   77.3  & 70.5     &  74.3     &   84.1     \\
GLASS~\cite{ronen2022glass} &    $-$     &  $-$     &   84.7     &   80.1   &  76.3   &   76.6     &  83.0       \\
\midrule
Ours &    \textbf{64.4}     &  \textbf{82.3}     &   84.8     &   \textbf{83.7}   &  \textbf{79.6}   &   \textbf{79.4}     &  \textbf{85.1}        \\
                         \bottomrule
\end{tabular}
}
    \vspace*{-1.00\baselineskip}
\end{table}

\subsection{Dataset and Evaluation}
We evaluate the end-to-end text spotting accuracy of the proposed method on several standard benchmarks.
We follow the standard evaluation protocols, which are based on F-measure evaluation~\cite{liu2019ctw1500, karatzas2015icdar, ch2020totaltext}.
The benchmark datasets used for the experiments in this paper are briefly introduced below.

\noindent
\textbf{CTW1500~\cite{liu2019ctw1500}:} is a curved scene benchmark, with 1,000 images for training and 500 images for testing.

\noindent
\textbf{ICDAR 2015~\cite{karatzas2015icdar}:} is collected as scene text images containing low resolution and containing small text instances.
It contains 1,000 training and 500 testing images.

\noindent
\textbf{Total-Text~\cite{ch2020totaltext}:} is another curved text benchmark, which consists of 1,255 training images and 300 testing images with multiple orientations.

\subsection{Implementation Details}

We follow the implementation procedure of ABCNet v2~\cite{liu2021abcnetv2}.
The backbone of the network is based on ResNet-50-FPN~\cite{lin2017fpn}.
For the dataset,
The model is pre-trained on a mixture of 150k synthesized data~\cite{liao2020masktextspotterv3}, 7k MLT data~\cite{nayef2019icdarmlt}, and the corresponding training data of each dataset. 
The pre-trained model is then fine-tuned on the training set of the target dataset.
We use the BERT base model (uncased) pre-trained on Wikipedia provided by hugging face as a pre-trained word embedding model.
We optimize our model using SGD with an image batch size of 8 on 4 A100 GPUs.
We train the model until 260k iterations with the initial learning rate of $10^{-2}$, which reduces to $10^{-3}$ and $10^{-4}$ at the 160k-th and 220k-th iteration, respectively.
Note that We set the hyper parameters to be $\alpha = 1$, $\beta = 1$, $\gamma = 0.6$ in experiments.

\subsection{Main Results}
In this section, we compare the performances of the proposed model with state-of-the-art methods on public datasets.
As shown in Table~\ref{tab:method_overall_result}, we have summarized various text spotting methods.
We see that A3S surpasses the recent competitive methods~\cite{ronen2022glass,liu2021abcnetv2, zhang2022testr, huang2022swintextspotter} on most of the metrics.

In CTW1500, our method achieves 64.4 and 82.3 for ``None'' and ``Full'', respectively.
In particular, the accuracy on ``None'', accuracy without lexicons, is significantly improved by 6.9 points compared to the baseline ABCNet v2.
This indicates that A3S is more effective when lexicons are unavailable.

In ICDAR2015 and Total-Text, the proposed method reaches competitive accuracy.
Similar to CTW1500, we confirm that A3S performs well when unavailable lexicons.
The proposed method performs better than GLASS~\cite{ronen2022glass}, which enhances features before text recognition based on the global information in an input image.
It is suggested that not only visual information but also semantic one is effective in scene-text spotting.

\begin{figure}[tb]
    \centering
    \begin{subfigure}{.43\textwidth}
        \includegraphics[width=\linewidth]{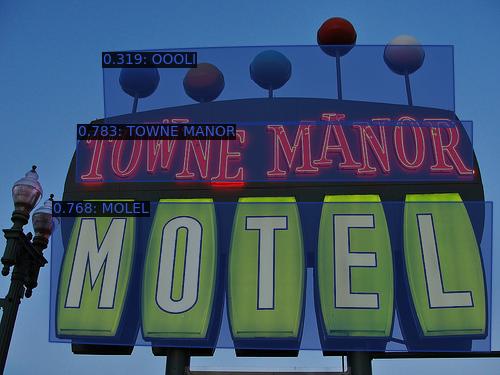}
        \caption{Baseline (ABCNet v2~\cite{liu2021abcnetv2})}
        \label{fig:result_abcnet}
    \end{subfigure}\hfill \\
    \begin{subfigure}{.43\textwidth}
        \includegraphics[width=\linewidth]{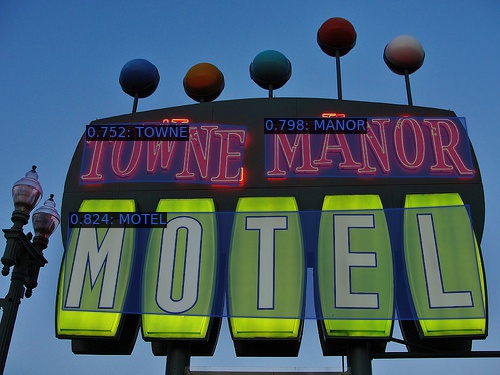}
        \caption{Our proposed method (A3S)}
        \label{fig:result_ours}
    \end{subfigure}\hfill
    \caption{
    Visualized end-to-end text spotting results of the proposed method and baseline on CTW1500.
    The detected area is shown in blue, and the predicted text and its confidence score are shown in the upper left corner of the detected area.
We see that the proposed method reduces both detection and text recognition errors.
    }
  \label{fig.detection_results}
    \vspace*{-1.00\baselineskip}
\end{figure}

We visualize the scene-text spotting result of ours and baseline (ABCNet v2~\cite{liu2021abcnetv2}) in Fig.~\ref{fig.detection_results}.
We see an improvement in detection and recognition compared to the baseline.

\begin{table}[bt]
\centering
\caption{
	Analysis of adversarial learning for semantic representations on CTW1500.
}
\label{tab:ablation_loss_function}
\resizebox{0.8\columnwidth}{!}{%0
\begin{tabular}{c|c|cc}
\toprule
Model                                & Method & None    & Full\\ \midrule
ABCNet v2 (baseline)~\cite{liu2021abcnetv2} & $-$    &  57.5  & 77.2 \\
model (a) & L1-Norm &  58.3  & 79.4 \\
model (b) & L2-Norm &  59.5  & 79.9 \\
\midrule
complete model & Adversarial learning &\textbf{64.4}  & \textbf{82.3} \\
\bottomrule
\end{tabular}
}
\vspace*{-1.00\baselineskip}
\end{table}

\subsection{Ablation Study}
To confirm the effectiveness of the proposed method in detail, we conducted ablation studies on CTW1500.

\noindent
\textbf{Impact of Adversarial Learning.}
We investigated the impact of adversarial learning on semantic representations.
Table~\ref{tab:ablation_loss_function} shows the accuracy results for the baseline, the complete proposed method, and the proposed model with different loss functions to match visual and semantic representations.
Models (a) and (b) show the models where the loss function is replaced by L1-Norm and L2-Norm, respectively.
Methods with semantic representations show accuracy improvement, but adversarial learning has a significant effect.
Since L1- and L2-Norm directly join different representations of images and language, the influence from the pre-trained language model is likely too significant.
Conversely, adversarial learning matches both representations indirectly by making them similar and thus is expected to be flexible and optimal.

\begin{table}[bt]
\centering
\caption{
	Impact of pre-trained word embedding on CTW1500.
	For GloVe, we utilized ``glove-wiki-gigaword-300'' from gensim.
}
\label{tab:ablation_word_embedding}
\resizebox{0.8\columnwidth}{!}{%0
\begin{tabular}{c|c|cc}
\toprule
Model                                & Word embedding & None    & Full\\ \midrule
model (c) & GloVe~\cite{pennington2014glove} &  61.1  & 80.2 \\
complete model & BERT~\cite{devlin2019bert} &\textbf{64.4}  & \textbf{82.3} \\
\bottomrule
\end{tabular}
}
\vspace*{-1.00\baselineskip}
\end{table}

\noindent
\textbf{Effect of Pre-trained Word Embedding.}
We investigated the impact of the pre-trained word embedding method in A3S.
The proposed method uses the pre-trained BERT~\cite{devlin2019bert}.
We examine the effect of using another word embedding model, GloVe~\cite{pennington2014glove}, which learns embedding based on global word-word co-occurrence statistics from a corpus. 
Table~\ref{tab:ablation_word_embedding} shows the difference in accuracy depending on the language model.
The model with GloVe is denoted as the model (c).
In both cases, we can see that using the language model improves accuracy compared to the baseline in Table~\ref{tab:ablation_loss_function}.
We see that the context-aware embedding method, BERT, has better accuracy, confirming the effectiveness of word embedding that utilizes context in scene-text spotting.

\section{Conclusion}\label{sec:conclusion}
In this paper, we propose a novel scene-text spotting method named A3S.
This method utilizes not only the visual information in text recognition but also the semantic one from a language model in training.
We proposed to leverage adversarial learning to connect visual and semantic representation flexibly.
Experiments show that our method achieves consistent gain and competitive results on popular benchmarks.

%\bibliographystyle{IEEEtran}
%\clearpage
{\small
\bibliographystyle{IEEEbib}
\bibliography{article}
}
%\bibliography{strings,refs}

\end{document}